%% file: acl_latex.tex
\pdfoutput=1

\documentclass[11pt]{article}

\usepackage{acl}
\usepackage{times}
\usepackage{latexsym}
\usepackage{graphicx}
\usepackage{algorithm}
\usepackage{algorithmic}
\usepackage{rotating}
\usepackage{balance}
\usepackage{color} 
\usepackage{soul}
\usepackage{xcolor}
\usepackage{colortbl} 

\usepackage{tabularx}
\usepackage{subfigure}
\usepackage{bm}
\usepackage{multirow}
\usepackage{enumitem}
\usepackage{booktabs}
\usepackage{amssymb}
\usepackage{amsmath}
\usepackage{colortbl}

\usepackage[T1]{fontenc}

\usepackage[utf8]{inputenc}

\usepackage{microtype}

\soulregister{\cite}7 
\soulregister{\citep}7 
\soulregister{\citet}7 
\soulregister{\ref}7 
\soulregister{\pageref}7 
\title{Improving Fake News Detection of Influential Domain \\via Domain- and Instance-Level Transfer}

\author{Qiong Nan\textsuperscript{1,2}, Danding Wang\textsuperscript{1}, Yongchun Zhu\textsuperscript{1,2}, Qiang Sheng\textsuperscript{1,2}  \\
{\bf Yuhui Shi\textsuperscript{1,2}, Juan Cao\textsuperscript{1,2}\thanks{~~Corresponding author}, Jintao Li\textsuperscript{1} }\\
        \textsuperscript{1}Key Laboratory of Intelligent Information Processing, \\ 
        Institute of Computing Technology, Chinese Academy of Sciences \\
        \textsuperscript{2}University of Chinese Academy of Sciences \\
        \texttt{\{nanqiong19z,wangdanding,zhuyongchun18s\}@ict.ac.cn} \\
        \texttt{\{shengqiang18z,caojuan,jtli\}@ict.ac.cn} \\
        \texttt{shiyuhui221@mails.ucas.ac.cn}}

\begin{document}
\maketitle
\begin{abstract}
Both real and fake news in various domains, such as politics, health, and entertainment are spread via online social media every day, necessitating fake news detection for multiple domains. Among them, fake news in specific domains like politics and health has more serious potential negative impacts on the real world (e.g., the infodemic led by COVID-19 misinformation).
Previous studies focus on multi-domain fake news detection, by equally mining and modeling the correlation between domains.
However, these multi-domain methods suffer from a seesaw problem: the performance of some domains is often improved at the cost of hurting the performance of other domains, which could lead to an unsatisfying performance in specific domains.
To address this issue, we propose a Domain- and Instance-level Transfer Framework for Fake News Detection (DITFEND), which could improve the performance of specific target domains. To transfer coarse-grained domain-level knowledge, we train a general model with data of all domains from the meta-learning perspective. To transfer fine-grained instance-level knowledge and adapt the general model to a target domain, we train a language model on the target domain to evaluate the transferability of each data instance in source domains and re-weigh each instance's contribution. Offline experiments on two datasets demonstrate the effectiveness of DITFEND. Online experiments show that DITFEND brings additional improvements over the base models in a real-world scenario.
\end{abstract}

\input{Introduction}
\input{Relatedwork}
\input{Model}
\input{Experiment}
\input{Conclusion}
\section*{Acknowledgements}
The research work is supported by the National Key Research and Development Program of China (2021AAA0140203), the National Natural Science Foundation of China (62203425), the China Postdoctoral Science Foundation (2022TQ0344), and the International Postdoctoral Exchange Fellowship Program by Office of China Postdoc Council (YJ20220198).
\bibliography{anthology}
\bibliographystyle{acl_natbib}

\appendix
\input{Appendix}

\end{document}

%% file: Introduction.tex
\section{Introduction}
With the rapid popularization of the Internet, more and more people tend to acquire news through social media platforms, such as Weibo\footnote{\url{https://weibo.com}} and Twitter\footnote{\url{https://twitter.com}}. 
Due to the above phenomenon, fake news has spread widely all over the world.
During the 2016 U.S. presidential election campaign, the top 20 frequently discussed fake election stories generated 8,711,000 engagements on Facebook, which is larger than the total of 7,367,000 for the top 20 most-discussed election stories posted by 19 major news websites~\cite{2016president}.
The wide spread of fake news may break the authenticity balance of the news ecosystem~\cite{shu2017fake},
and it not only misleads many people but also leads to the social mobs and social panic~\cite{chinese_rumor}.
Therefore, fake news detection is of significant importance.

\begin{figure}[t]
\setlength{\abovecaptionskip}{3pt}
\setlength{\belowcaptionskip}{-11pt}
\centering
\begin{minipage}[b]{1\linewidth}
\centering
\includegraphics[width=1.0\linewidth]{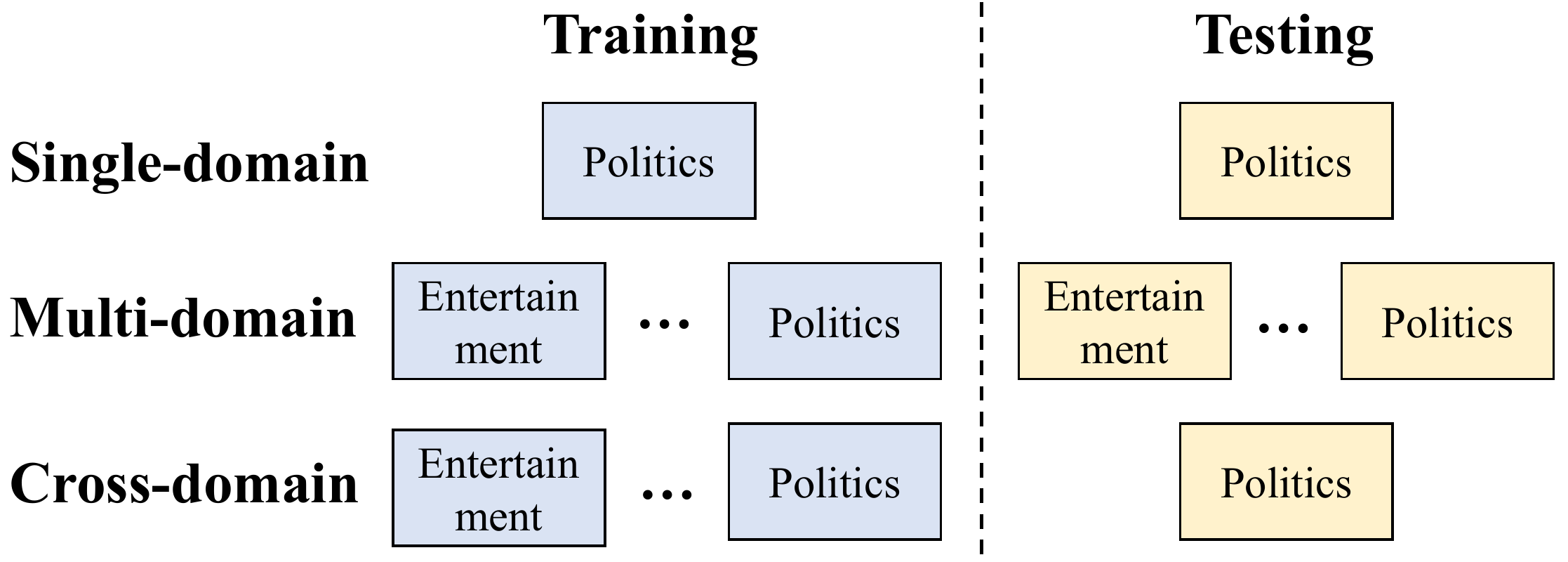}
\end{minipage}
\caption{Illustration of the differences among single-domain, multi-domain, and cross-domain (\textbf{ours}) fake news detection.}\label{fig:intro}
\end{figure} 

Real-world news platforms categorize news pieces by topic into various domains, e.g., politics, health, and entertainment. 
Although it is necessary to detect fake news in every domain, the societal effects of fake news from some specific domains are more serious. For example,
during the U.S. election, fake news in the political domain may have dictated election results~\cite{allcott2017social}, and
\citet{science18} find that
false political news travels deeper more quickly and more broadly, reaches more people, and is more vital than other categories of false information.
In the COVID-19 infodemic~\cite{bursztyn2020misinformation}, thousands of fake news pieces have caused social panic~\cite{chinese_rumor} and weakened the effect of pandemic countermeasures~\cite{bursztyn2020misinformation}. 
Therefore, it is important to detect fake news for these specific news domains, e.g., politics~\cite{wang2017liar,jin2017detection}, health~\cite{bang2021model,shi2020rumor}. Some researchers pay attention to fake news detection in the influential news domains by modeling each domain separately~\cite{wang2017liar,allcott2017social,bovet2019influence,dai2020ginger,cui2020deterrent,zhou2020recovery, shang2022duo}. However, a 
single-domain dataset could only contain limited data, e.g., the PolitiFact dataset of the political domain only has 948 samples~\cite{shu2020fakenewsnet}.
Fortunately, news pieces in different domains are correlated~\cite{nan2021mdfend,silva2021embracing}, so it is promising to exploit data of other domains to improve performance for a target domain.

Recently, some works~\cite{wang2018eann,silva2021embracing, nan2021mdfend} simultaneously model various domains to improve the overall performance of all domains.
However, these multi-domain methods suffer from a serious seesaw phenomenon which could cause the performance of some domains to be improved at the cost of hurting the performance of other domains~\cite{tang2020progressive}. For example, EANN~\cite{wang2018eann} performs quite well in \emph{Military} (with f1-score of 0.9274), whereas the results is not satisfactory in \emph{Politics} (with f1-score of 0.8705)~\cite{nan2021mdfend}.
Moreover, it is hard to guide these multi-domain models to improve the performance of a specific target domain due to the lack of a target-oriented design.
In this paper, we focus on exploiting news pieces of other domains to improve the detection performance of a certain target domain, called \textbf{cross-domain fake news detection}, which can bring additional gains for target domains compared to multi-domain methods. 
The differences among single-domain, multi-domain, and cross-domain fake news detection are shown in Figure~\ref{fig:intro}. To solve cross-domain fake news detection, we adopt two key ideas:

\textbf{Transfer domain-level knowledge.} It is necessary to transfer knowledge from multiple domains because news pieces in different domains are correlated. Moreover, to alleviate the seesaw phenomenon, improving the generalization ability of a multi-domain model, which can adapt fast to the target domain is necessary. 

\textbf{Transfer instance-level knowledge.} The transferability varies from instance to instance. For example, ``\emph{A politician claimed COVID-19 is less lethal than flu}'' and ``\emph{New York Officials welcome immigrants, legal or illegal}'' are two different news pieces of politics. The former is more relevant to health while the latter is irrelevant. In other words, the former is more transferable. Therefore, it is important to quantify the transferability of source instances, in order to decrease the impact of irrelevant instances.

Along this line, we propose a \underline{D}omain- and \underline{I}nstance-level \underline{T}ransfer framework for \underline{F}ak\underline{e} \underline{N}ews \underline{D}etection (DITFEND). To transfer domain-level knowledge, we exploit data from multiple domains to train a general model from the meta-learning perspective, which contains common knowledge and can adapt fast to a specific domain. To transfer instance-level knowledge, we first learn a domain-adaptive language model which is endowed with characteristics of the target domain. 
To weigh the contribution to the target domain of every instance in source domains, we adopt an index, \textbf{perplexity} of the domain-adaptive language model, to quantify the transferability of these instances. Because in information theory perplexity can measure how well a probability model predicts an instance, in other words, low perplexity on a news piece indicates that the instance is highly related to the knowledge contained in the language model.
Finally, we adapt the general model with instances from the target domain and weighted instances from source domains, in order to achieve satisfying performance on the target domain.

The main contributions of this paper can be summarized as follows:
\begin{itemize}
    \setlength{\itemsep}{0 pt}
    \item We investigate the importance of cross-domain fake news detection with multiple sources for target domains for the first time.
    \item {We propose a Domain- and Instance-level Transfer Framework to improve fake news detection of target domains.}
    \item {We evaluate our proposed DITFEND on both English and Chinese real-word fake news datasets, and experiments demonstrate the effectiveness of DITFEND.}
\end{itemize}

%% file: Relatedwork.tex
\section{Related Work}
\subsection{Fake News Detection}
Fake news detection aims at automatically classifying a news piece as real or fake.
Existing methods can be generally grouped into two clusters: social-context-based methods and content-based methods~\cite{shu2017fake}. For social-context-based methods, some analyze propagation patterns to mine structural signals for fake news detection~\cite{jin2014news, liu2018early, shu2019defend, mosallanezhad2022domain, naumzik2022detecting}, others use the wisdom of crowds, such as emotion and stance, to detect fake news~\cite{zhang2021mining,jin2016news}, and \citet{sheng-etal-2022-zoom} captures the environmental signals to detect fake news posts. For content-based methods, some extract evidence from external sources~\cite{vlachos2014fact, shi2016fact, chen2022evidencenet, xu2022evidence,sheng2021integrating}, while others only analyze news itself and focus on better constructing features~\cite{przybyla2020capturing, wang2018eann}, which is within the scope of our research.

Since real-world news platforms categorize news pieces by topic into various domains, some researchers pay attention to the fake news detection performance of each domain, especially for that with serious societal effects. Some only take one specific domain into consideration and perform single-domain fake news detection~\cite{wang2017liar,allcott2017social,bovet2019influence,dai2020ginger,cui2020deterrent,zhou2020recovery, shang2022duo}, however, they ignore useful information from other domains. Some methods simultaneously model various domains to improve the overall performance of all domains (multi-domain fake news detection)~\cite{wang2018eann,silva2021embracing,nan2021mdfend, zhu2022memory}, however, due to the seesaw phenomenon, the performance on some target domains suffer from degeneration. \citet{huang2021dafd} and \citet{mosallanezhad2022domain} propose to use a domain adaptation strategy for cross-domain fake news detection, however, they only transfer knowledge from one single source domain to ensure the model's detection performance on the target domain. However, in the practical scenario, news pieces from multiple source domains are inherently correlated with the target domain. It is straightforward to combine all source domains into one single domain and perform cross-domain fake news detection as  \cite{huang2021dafd} and \cite{mosallanezhad2022domain} do, but the improvement may not be significant. Hence, it is necessary to find a better way to make full use of all source domains to improve the performance of a target domain.

\subsection{Transfer Learning}
Transfer learning aims to leverage knowledge from a source domain to improve the learning performance or minimize the number of labeled examples required in a target domain~\cite{pan2009survey,zhuang2020comprehensive}. Recently, transfer learning has been widely adopted for natural language processing~\cite{devlin2019bert,liu2019roberta,lewis2020bart}, e.g., sentiment classification~\cite{peng2018cross}, neural machine translation~\cite{kim2019effective}, style transfer~\cite{yang2018unsupervised}. 
Meanwhile, meta learning serves as a paradigm that can be used to improve transfer learning problems~\cite{hospedales2021meta}.
Benefit from its ability to integrate prior experience as well as generality to all domains, meta learning has been widely adopted in many applications
~\cite{li2018learning, wang2021multimodal}. 
In this paper, we propose a novel transfer framework based on meta-learning for cross-domain fake news detection.

%% file: Model.tex
\section{DITFEND: Domain- and Instance- Level Transfer for Fake News Detection}
\begin{figure}
	\centering
	\includegraphics[scale = 0.4]{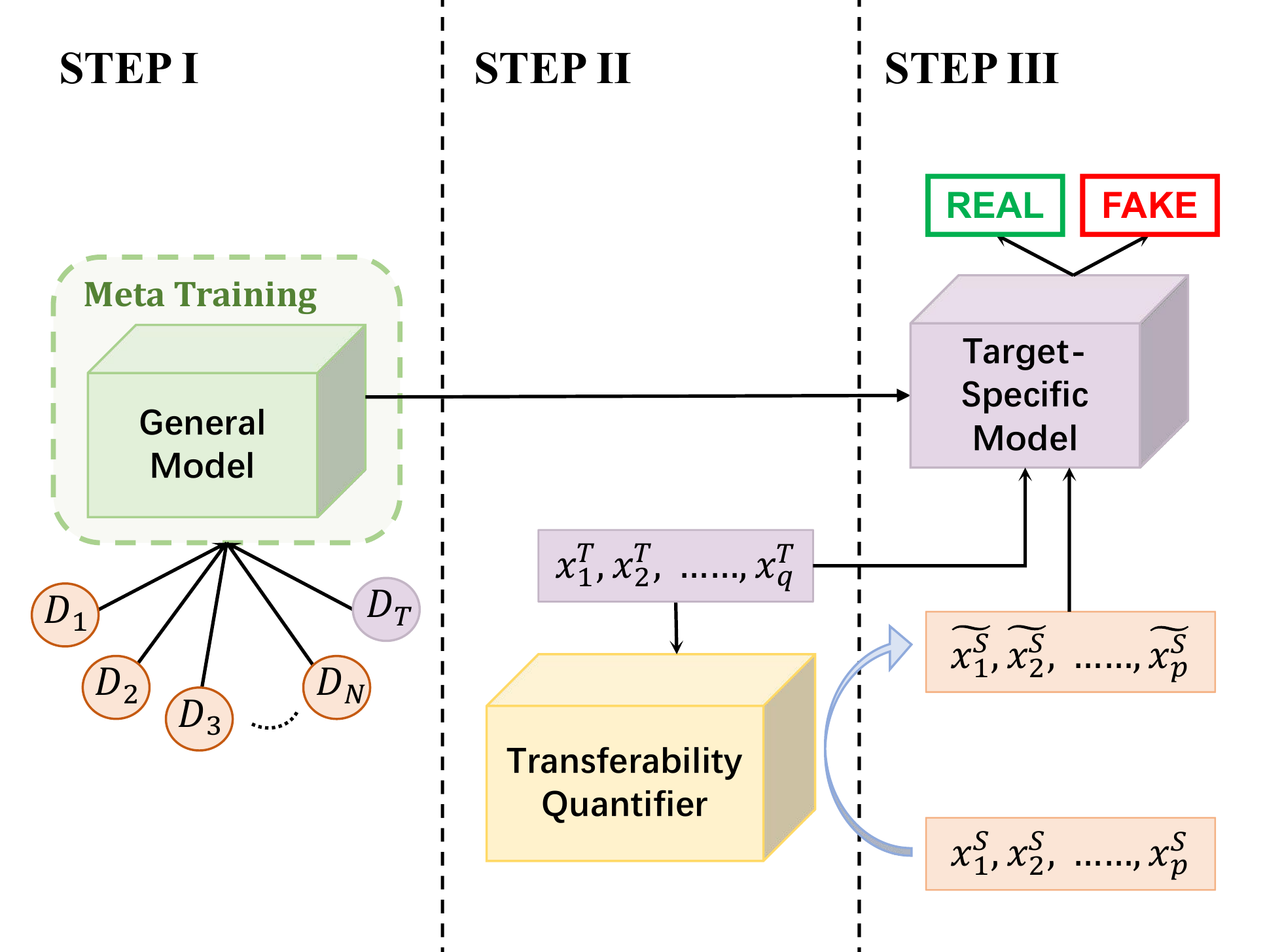}
	\caption{The overall framework of our DITFEND model. Step I is the general model training procedure, step II is the transferability quantifying procedure, and step III is the target domain adaptation procedure. The three steps correspond to Section \ref{general}, \ref{transferability} and \ref{adaptation} respectively.}
	\label{fig:model}
\end{figure}
\subsection{Problem Statement}
In the scenario of cross-domain fake news detection, there is one target domain $\mathcal{D}_T$ and a collection of $N$ source domains $\mathcal{D}_S = \{ \mathcal{D}_{1}, \mathcal{D}_{2},... ,\mathcal{D}_{N}\}$. For each domain, we have a collection of news pieces.  
For a piece of news $P$,
we pad it with additional characters $\scriptstyle [CLS]$ and $\scriptstyle [SEP]$. 
Then we tokenize the sentence into $m$ tokens $\{t_1, t_2, ..., t_m\}$,
and encode it to get the corresponding embedding vector \bm{$e$}. To distinguish between different domains, we denote \bm{$e_T$} as the embedding vector of the target domain and $\bm{e_S} \in \left\{\bm{e_1}, \bm{e_2}, ..., \bm{e_N}\right\}$ of source domains.

The core idea of our framework is to leverage news pieces from all domains to train a target-adaptive fake news detector.
To achieve this, in the first stage, we train a general model with news pieces from all domains, such that the model can alleviate the seesaw phenomenon (i.e., domain-level transfer); In the second stage, we evaluate and quantify the transferability of instances from source domains; And in the third stage, we adapt the general model to the target domain. The overall framework is shown in Figure \ref{fig:model}. 
\subsection{General Model Training}
\label{general}
\begin{algorithm} [t] 
	\small
	\caption{General Model Training Procedure.}
	\label{alg: general model training}
	\flushleft{\textbf{Input}: Given $N$ source domains and one target domain.
		\textbf{Input}: The base model $\mathcal{M}(\theta)$. \\
		\textbf{Input}: Learning rate $\alpha$ and $\beta$.\vspace{-0.2cm}
		\begin{enumerate}
			\setlength{\itemsep}{0pt}
			\setlength{\parsep}{0pt}
			\setlength{\parskip}{2pt}
			\item[1.] randomly initialize $\theta$
			\item[2.] \textbf{while} not coverage \textbf{do}:
			\item[3.] \qquad sample batch of training tasks $\{\mathcal{T}_1, ..., \mathcal{T}_n\}$
			\item[4.] \qquad \textbf{for} all $\mathcal{T}_d \in \{\mathcal{T}_1, ..., \mathcal{T}_n\}$ \textbf{do}:
			\item[5.] \qquad \qquad $\mathcal{T}_d$ contains two disjoint sets $\mathcal{D}_d^s$ and $\mathcal{D}_d^q$
			\item[6.] \qquad \qquad evaluate loss $\mathcal{L}_d^s(\theta)$ with $\mathcal{D}_d^s$
			\item[7.] \qquad \qquad compute updated parameter\\ \qquad \qquad $\theta_d = \theta - \alpha \frac{\partial \mathcal{L}_d^s(\theta)}{\partial \theta}$
			\item[8.] \qquad \qquad evaluate loss $\mathcal{L}_d^q(\theta_d)$ with $\mathcal{D}_d^q$
			\item[9.] \qquad \textbf{end}
			\item [10.] \qquad update $\theta = \theta - \beta \sum_{\mathcal{T}_d \in \{\mathcal{T}_1, ...,\mathcal{T}_n\}} \frac{\partial \mathcal{L}_d^q(\theta_d)}{\partial \theta}$
			\item[11.] \textbf{end}
		\end{enumerate}
		\vspace{-0.2cm}
	}
\end{algorithm}
For domain-level transfer, we take the advantage of meta-learning and train a general model $\mathcal{M}(\theta)$ to aggregate the knowledge from all the domains.

In each iteration of the parameter update, we draw a batch of training tasks $ \left\{\mathcal{T}_{1},... ,\mathcal{T}_{n}\right\}$, and for each task $\mathcal{T}_d \in \left\{\mathcal{T}_{1},... ,\mathcal{T}_{n}\right\}$,
we divide it into two disjoint sets: a support set $\mathcal{D}_d^s$ and a query set $\mathcal{D}_d^q$.
The model is trained with samples from support set $\mathcal{D}_d^s$ and feedback with the corresponding loss $\mathcal{L}_d^s$ from $\mathcal{D}_d^s$:
\begin{equation}
	\mathcal{L}_d^{s}(\theta) = \frac{1}{m_s}\sum_{i = 1}^{m_s}{-y_i\log\hat{y_i} - (1 - y_i)\log(1 - \hat{y_i})},
\end{equation}
where $m_s$ is the number of the data in the current support set. We adopt cross-entropy loss because it is widely used as the optimization goal for binary fake news detection~\cite{wang2018eann, silva2021embracing, nan2021mdfend}. And then we use $\mathcal{L}_d^s$ to optimize the parameter of the current task $\mathcal{T}_d$ via gradient descent:
\begin{equation}
	\theta_d = \theta - \alpha \frac{\partial \mathcal{L}_d^s(\theta)}{\partial \theta},
\end{equation}
where $\alpha$ is the learning rate of the training process within each task, and $\theta_d$ is the optimized model parameters for the current task $\mathcal{T}_d$.

And then the model is tested on samples from the current query set $\mathcal{D}_d^q$. The model is improved by considering how the test error $\mathcal{L}_d^{q}$ on $\mathcal{D}_d^q$ changes with respect to the parameters:
\begin{equation}
    \begin{split}
	\mathcal{L}_d^{q}(\theta_d) = \frac{1}{m_q}\sum_{i = 1}^{m_q}&-y_i\log\hat{y_i} \\
 &- (1 - y_i)\log(1 - \hat{y_i})
    \end{split}
\end{equation}
where $m_q$ is the number of data in the current query set.

In effect, the testing error on $\mathcal{D}_d^q$ serves as the training error of the meta-learning process.

After we loop over all tasks in $\left\{\mathcal{T}_{1},... ,\mathcal{T}_{n}\right\}$, 
the base model's parameters $\theta$ can be updated as follows:
\begin{equation}
	\theta = \theta - \beta \nabla_\theta\sum_{i = 1}^{n}\mathcal{L}_i^q,
\end{equation}
where $\beta$ is the learning rate of the meta-learning process.
The procedure of general model training is summarized in Alg~\ref{alg: general model training}.

\subsection{Transferability Quantifying}
\label{transferability}
\begin{figure*}
    \centerline{\includegraphics[scale = 0.38]{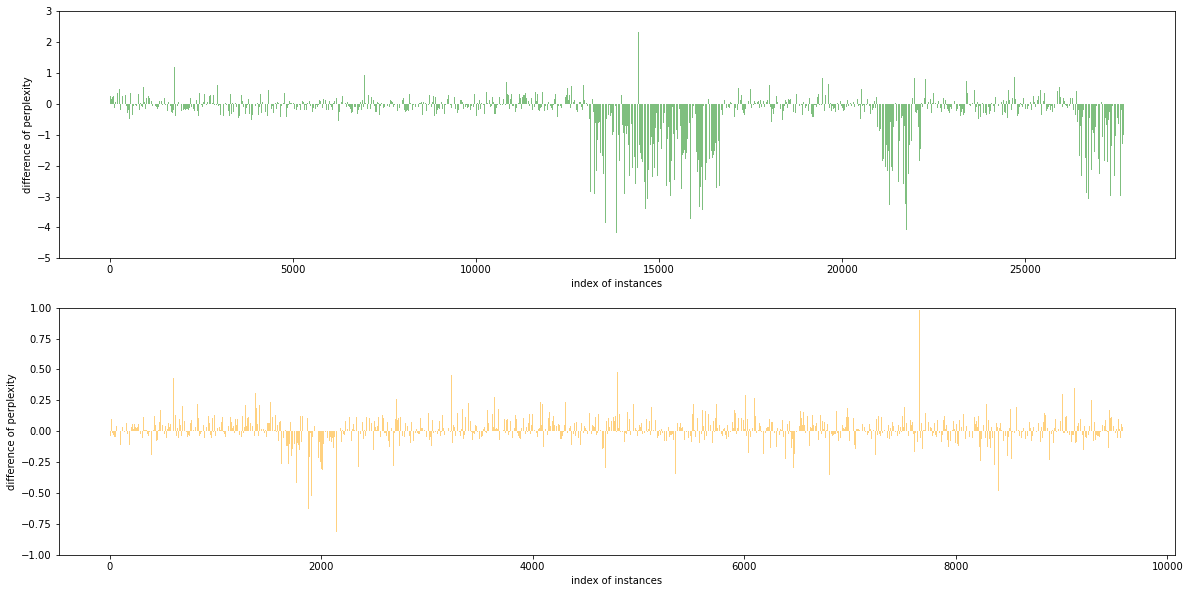}}
    \caption{The D-value of different target language models' perplexity on the same batch of source instances (Top: Chinese dataset; Bottom: English dataset). }
    \label{fig: pp distribution}
\end{figure*}
In order to endow the language model with target domain knowledge, we perform the Masked Language Modeling task on the target domain to get a domain-adaptive language model. This second phase of training the language model can bring significant performance improvement on the following task based on the language model~\cite{gururangan2020don}.
And we utilize the domain-adaptive language model to evaluate the transferability of source instances.

\textbf{Domain-adaptive Language Model Training.} Let $\mathcal{D}_T$ be a dataset of news pieces from the target domain $T$, where $P_T$ is a news piece in $\mathcal{D}_T$ containing $n$ tokens $\left\{w_1, ..., w_n\right\}$. We replace 15\% tokens in the input sequence with the $\scriptstyle [MASK]$ token, a random token, or the original token. The model is required to predict the masked tokens based on the other tokens in $P_T$. The training objective is to minimize the cross-entropy loss of the language model in predicting the masked tokens of the target data:
\begin{equation}
	\min_{\theta_{m}}\sum_{x\in D_t}\mathcal{L}_{m}(x, \theta_{m}),
\end{equation}
where $\theta_{m}$ is the parameters of the language model (\emph{bert-base-uncased} for English and \emph{bert-base-chinese} for Chinese).

After performing the Masked Language Modeling task on the target domain, we can effectively endow the corresponding language model with knowledge of the target domain.

\textbf{Perplexity} is an indicator that can measure how well a language model predicts a sentence, i.e., the higher the probability of prediction is, the better the language model is, and the lower the language model's perplexity is. Many researchers use it to evaluate language models~\cite{belinkov2019analysis,yogatama2018memory}. We exploit perplexity from another perspective -- the better a language model can predict a sentence, the better this sentence fits the knowledge endowed in the language model. Therefore, it is intuitive to compute the perplexity of the language model on a given sample to quantify the sample's transferability, i.e., the lower perplexity indicates stronger transferability. 

Let $\mathcal{D}_S$ indicate one of the source datasets, where $P_S$ is a news piece in $\mathcal{D}_S$ containing $m$ tokens $\left\{w_0, w_2, ..., w_{m-1}\right\}$. We pad it with two tokens, i.e., $\scriptstyle [CLS]$ and $\scriptstyle [SEP]$. A masked sentence $P_{mask}$ is generated by masking a word/character in the sentence as follows:
\begin{equation}
	\mathcal{P}_{mask} = <[CLS], ..., w_{i-1}, \small{[MASK]}, w_{i+1}, ..., [SEP]>,
\end{equation}
where $0\leq i \leq m-1$. Then the target-adaptive language model is utilized to predict the probability of the right words $w_i$ in the $\scriptstyle [MASK]$ position:
\begin{equation}
	prob(w_i) = MLM(\mathcal{P}, w_i).
\end{equation}

After we calculated the probability of all the words in the sentence, the perplexity of the language model on the whole sentence is calculated as follows:
\begin{equation}
	pp = \sqrt[N]{\prod \limits_{i=0}^{N-1}\frac{1}{prob(w_i)}}.
\end{equation}

To show the difference in the target-adaptive language model when assigning different target domains, we compute the difference value of different target-adaptive language models' perplexity on source instances. And we visualize the distribution in Figure \ref{fig: pp distribution} in detail. From the distribution, we can see that there exists an obvious difference in both the Chinese and English datasets.

Finally, we quantify the transferability of instances from source domains as follows:
\begin{equation}
	w = 1 / pp,
\end{equation}
where $w$ is the indicator of the transferability of the sample. In this way, we quantify the transferability of source instances, i.e., assigning bigger weights to samples with lower perplexity. 
\subsection{Target Domain Adaptation}
\label{adaptation}
In this section, we aim to adapt the general fake news detection model to the target domain. In Section \ref{transferability}, we have assigned each sample in source domains with a weight value to indicate its transferability. To make full use of source domain samples, we re-weigh them based on their transferability, along with the target domain samples to train the general model.

We exploit the general model trained via meta-learning (Section \ref{general}), and optimize it by minimizing the cross-entropy loss as follows:
\begin{equation}
	\begin{aligned}
		\mathcal{L}^{ce}(y, \hat{y}) &= -ylog\hat{y} - (1-y)log(1-\hat{y}) \\
		\mathcal{L} &= \mathbb{E}_{(x,y) \sim p_s(x,y)} w(x)\mathcal{L}^{ce}(y, \hat{y}) \\
		&+\mathbb{E}_{(x,y) \sim p_t(x,y)}\mathcal{L}^{ce}(y, \hat{y}),
	\end{aligned}
\end{equation}
where $y$ is the ground truth, $\hat{y}$ is the predicted label, and $w(x)$ is the indicator of the transferability of the instance $x$ obtained in Section \ref{transferability}.

%% file: Experiment.tex
\section{Experiments}
\label{sec:experiment}
In this section, we aim to answer the following evaluation questions:
\begin{itemize}
    \setlength{\itemsep}{0 pt}
    \item \textbf{EQ1:} Can DITFEND improve the performance of a target domain when coordinates with different base models?
    \item \textbf{EQ2:} How effective are domain-level transfer and instance-level transfer?
    \item \textbf{EQ3:} Can DITFEND perform well in a real-world fake news detection scenario?
\end{itemize}
\subsection{Datasets}
We evaluate DITFEND on both Chinese and English datasets. And the statistics of datasets are listed in Tables \ref{tab:chinese} and \ref{tab:english}.
\input{tables/chinese_dataset}
\input{tables/english_dataset}

\textbf{English Dataset.} We combine two datasets (PolitiFact and GossipCop) in FakeNewsNet~\cite{shu2020fakenewsnet} and COVID~\cite{li2020mm} into an English dataset, which contains three domains, namely PolitiFact, GossipCop and COVID.

\textbf{Chinese Dataset}~\cite{nan2021mdfend}.  multi-domain fake news detection dataset collected from Sina Weibo. There are 9 domains in total, which is Science, Military, Education, Disaster, Politics, Health, Finance, Entertainment, and Society. In our experiments, in order to evaluate the impact of the domain number, we sample 3 domains (Politics, Health, Entertainment) and 6 domains (Education, Disaster, Health, Finance, Entertainment, Society) from the Chinese dataset to construct the other two multi-domain datasets, i.e., Chinese 3-domain and Chinese 6-domain. 

\input{tables/en_ch3}
\subsection{Baseline Models}
In this paper, we focus on textual content-based fake news detection. 
Technically, our DITFEND framework can coordinate with any text-based models that produce post-representation.
Thus, we use 8 representative text-based models as our base models, which can be divided into three groups: (1) text classification models: \textbf{TextCNN}~\cite{kim-2014-convolutional}, \textbf{BiGRU}~\cite{ma2016detecting}, and \textbf{RoBERTa}~\cite{liu2019roberta,cui2019pre}; (2) transfer learning models: \textbf{MMOE}~\cite{ma2018modeling} and \textbf{MOSE}~\cite{qin2020multitask}; (3) multi-domain fake news detection models: \textbf{EANN\scriptsize{T}}~\cite{wang2018eann}, \textbf{EDDFN}~\cite{silva2021embracing}, and \textbf{MDFEND}~\cite{nan2021mdfend}. Details about the base models are as follows:

\textbf{TextCNN}~\cite{kim-2014-convolutional}: Convolutional Neural Networks (CNNs) have been proven to gain remarkably strong performance on the task of text classification. In our experiment, we use filter windows of 1, 2, 3, 5, 10 with 64 feature maps each.

\textbf{BiGRU}~\cite{ma2016detecting}: It is a widely used models in natural language processing applications. Different from ~\cite{ma2016detecting}, we treat each piece of news as a sequential input to a one-layer BiGRU model.

\textbf{RoBERTa}~\cite{liu2019roberta,cui2019pre}: It is a robustly optimized BERT~\cite{devlin2019bert} pre-trained model. We utilize it to encode tokens of news content and feed the extracted embedding into an MLP to obtain the final prediction. \emph{roberta-base-chinese} and \emph{roberta-base-uncased}~\cite{liu2019roberta,cui2019pre} are exploited for Chinese and English datasets, respectively.

\textbf{EANN\scriptsize{T}}~\cite{wang2018eann}: It is a model that aims to learn event-agnostic features, which uses a TextCNN module for text representation and adopts an auxiliary event discriminator for adversarial learning. In our experiments, we only use the textual branch, and rather than event-level, we perform domain-level adversarial training (i.e., use the discriminator to classify different domains) to learn domain-shared features.

\textbf{MMOE}~\cite{ma2018modeling} and \textbf{MOSE}~\cite{qin2020multitask}: These two models are proposed for multi-task learning. In our experiments, we assume that fake news in different domains are different tasks, and use the two models for multi-domain modeling.

\textbf{EDDFN}~\cite{silva2021embracing}: It is proposed for multi-domain fake news detection, which models different domains implicitly and jointly preserves domain-specific and cross-domain knowledge.

\textbf{MDFEND}~\cite{nan2021mdfend}: It is a multi-domain fake news detection model, which utilizes a domain gate to aggregate multiple representations extracted by mixture-of-experts for multi-domain fake news detection.

\subsection{Experiment Settings}
\textbf{Training Procedure.} \textbf{I.} We use samples from all domains to train a general model from the meta-learning perspective. \textbf{II.} We use samples from the target domain to train a domain-adaptive language model via \emph{Masked Language Modeling} task. \textbf{III.} We calculate the perplexity of the domain-adaptive language model on samples from source domains, which is used to re-weigh the samples. \textbf{IV.} We use samples from the target domain and source domains (assigned with the corresponding weights) to train the general model to adapt to the target domain.

\textbf{Target domain assignment.} In practice, fake news pieces of Politics, Health, and Finance have more serious influence. For Ch-3 and Ch-9 dataset, we choose \emph{Politics} and \emph{Health} as the target domain, respectively. For Ch-6 dataset, we choose \emph{Health}  and \emph{Finance} as the target domain,respectively. For English dataset, we choose \emph{PolitiFact} and \emph{COVID} as the target domain, respectively.

\textbf{Evaluation Metrics.} 
We treat the fake news detection problem as a binary classification task. We report macro F1 score (F1), accuracy (Acc), and Area Under ROC (AUC). For the online tests, where the number of fake news is much lower than real news, we should detect fake news as accurately as possible without misclassifying real news. Thus, we further report standardized partial AUC (SPAUC$_{FPR \leq 0.1}$). 

\textbf{Implementation Details.} We limit the max length of the sentence to 170 tokens for Chinese, and 300 tokens for English. Following the settings in \cite{nan2021mdfend}, we tokenized sentences with jieba\footnote{https://github.com/fxsjy/
jieba} (for Chinese), nltk~\cite{nltk} (for English), and embed them with Word2Vec~\cite{le2014distributed, mikolov2013efficient} for BiGRU and TextCNN. We use RoBERTa and the corresponding tokenizer for other models. 
For the two embedding types, we fix the dimension of embeddings to 768 for RoBERTa~\cite{devlin2019bert} and 200 for Word2Vec~\cite{le2014distributed,mikolov2013efficient}. 
For a fair comparison, we set the same hyperparameters for all base models. The MLP (Multi-Layer Perceptron) module used in these models contains one dense layer (384 hidden units). 

\subsection{Performance Comparison (EQ1)}

To answer \textbf{EQ1}, the DITFEND framework coordinates with each base model and compares with it respectively on both the Chinese dataset and English dataset. Experiment results are shown in Table \ref{tab:ench3} and Table \ref{tab:ch6_ch9} (Appendix \ref{sec: ch69experiments}). The experimental results further reveal several insightful observations:
\begin{itemize}
    \setlength{\itemsep}{0 pt}
    \item DITFEND can coordinate with various base models. From the results compared with these base models, DITFEND can mostly bring improvements of the fake news detection performance on target domains, which indicates DITFEND has satisfying compatibility.
    \item The improvement on the English dataset is bigger than the Chinese dataset. The main reason could be that the English dataset contains more samples than Chinese dataset, which can bring benefits for knowledge transfer. During target domain adaptation, the model can learn more transferable information about the target domain.
    \item We find that for the same target domains (Health and Politics) in Ch-3 dataset and Ch-9 dataset, the performance of the target domain in Ch-9 dataset exceeds the former with most base models, while on the other hand, for the same target domain (Health) in Ch-6 and Ch-9, the performances are too close to call, 
    which indicates that more domains could bring more transferable knowledge to some extent, but the number of domain is not proportionate with performance. 
\end{itemize}


\subsection{Analysis (EQ2)}
To answer \textbf{EQ2}, we conduct both an ablation study and a case study.
In the ablation study, we evaluate the effectiveness of domain-level transfer and instance-level transfer respectively; In case study, we aim to intuitively illustrate that our transferability quantifying strategy can pick out related source news pieces according to one specific domain. 
\input{tables/ablation_ch9}
\subsubsection{Ablation Study} 
We evaluate two ablation experiment groups based on two representative fake news detection base models to evaluate the effectiveness of different modules in DITFEND framework. Table \ref{tab:abla_ch9} shows the experimental results. To evaluate the effectiveness of domain-level transfer, we replace the meta-training procedure with classical training and use the weighted source data and target data to train the general model to adapt to the target domain. From the results, we can see that $w/o$ \emph{meta-training} performs worse than the whole DITFEND framework but better than the original model. To evaluate the effectiveness of instance-level transfer, we abandon instances from source domains when performing domain adaptation. The ablation results show that without instances from source domains, performance on target domains drops a little bit. Comparing the degree of decline of the two ablation experiments, we can also conclude that meta-training procedure brings more boost compared to instance from source domains.

\subsubsection{Case Study}
To further verify whether our transferability quantifier can pick out useful samples from source domains effectively, we conduct a case study on both the Chinese dataset and the English dataset. For each dataset, we have two assigned target domains (i.e., \emph{Health} and \emph{Politics} for Chinese, \emph{COVID} and \emph{PolitiFact} for English). According to Section \ref{general}, each instance has two different transferability indicators depending on two different target domains. Therefore, we use the corresponding indicator to choose samples relevant to the target domain, which can illustrate the interpretability of instance-level transfer to some extent. Representative samples are listed in Table \ref{tab:case}. Taking \emph{Politics} domain (as the target domain) in the Chinese dataset for example, we successfully pick out some instances from other domains, which contain some knowledge related to \emph{Politics}.
\input{tables/pp_compare}
\subsection{Online Tests (EQ3)}
To verify the real benefits of DITFEND bringing to the online system, we conduct online testing experiments.
Different from the offline datasets, this online dataset is much more skewed (30,977 real: 774 fake $\approx$ 40:1).
The testing set contains all news pieces in a whole month (ranging from 2021/10/10 to 2021/11/10), and the training set contains news pieces published before 2021/10/10. Both the training set and the test set come from the same online system\footnote{http://www.newsverify.com/os/index.html}. 
The DITFEND coordinates with two online baselines: TextCNN and EANN, respectively. 
In real-world scenarios, the number of fake news items is much smaller than real news items, which means that we should detect as many fake news items as possible without misclassifying real news items. In other words, the objective is to improve the True Positive Rate (TPR) on the basis of a low False Positive Rate (FPR). Thus, beyond AUC and F1, we follow~\cite{mcclish1989analyzing,zhu2020modeling} and adopt standardized partial AUC (SPAUC$_{\text{FPR}\leq 0.1}$). The online results in Table~\ref{tab:online} demonstrate that DITFEND achieves a significant improvement on AUC and SPAUC against the baselines, with greater improvements in SPAUC.
\input{tables/online_result}

%% file: tables/chinese_dataset.tex

\begin{table}[htbp]
  \centering
  \caption{Statistics of the Chinese dataset}
    \scalebox{0.75}{
    \begin{tabular}{rrrrrr}
    \toprule
    \textbf{Domain} & \multicolumn{1}{c}{\textbf{Science}} & \multicolumn{1}{c}{\textbf{Military}} & \multicolumn{1}{c}{\textbf{Edu.}} & \multicolumn{1}{c}{\textbf{Disaster}} & \multicolumn{1}{c}{\textbf{Politics}} \\
    \midrule
    \# Fake  &  93    &   222    &   248    &    591   &  546\\
    \# Real  &  143    &   121    &   243    &   185    &  306\\
    Total &    236   &   343    &   491    &   776    &  852\\
    \midrule
    \textbf{Domain} & \multicolumn{1}{c}{\textbf{Health}} & \multicolumn{1}{c}{\textbf{Finance}} & \multicolumn{1}{c}{\textbf{Ent.}} & \multicolumn{1}{c}{\textbf{Society}} & \multicolumn{1}{c}{\textbf{All}} \\
    \midrule
    \# Fake  &  515   &   362    &   440    &   1,471    &  4,488\\
    \# Real  &  485   &    959   &   1,000    &   1,198    &  4,640\\
    Total &   1,000    &   1,321    &   1,440    &   2,669    &  9,128\\
    \bottomrule
    \end{tabular}%
    }
  \label{tab:chinese}%
\end{table}%

%% file: tables/english_dataset.tex
\begin{table}[htbp]
  \centering
  \caption{Statistics of the English Dataset}
    \scalebox{0.8}{
    \begin{tabular}{rrrrr}
    \toprule
    \textbf{Domain} & \multicolumn{1}{c}{\textbf{PolitiFact}} & \multicolumn{1}{c}{\textbf{GossipCop}} & \multicolumn{1}{c}{\textbf{COVID}} &
    \multicolumn{1}{c}{\textbf{All}}\\
    \midrule
    \# Fake  & 420 & 4,947 & 1,317 & 7,483\\
    \# Real  & 528 & 16,694 & 4,750 &  22,864\\
    Total & 948 & 21,641 & 6,067 &  30,347\\
    \bottomrule
    \end{tabular}%
    }
  \label{tab:english}%
\end{table}%

%% file: tables/en_ch3.tex
\begin{table*}[htbp]
  \centering
  \caption{Performance comparison of base models with and without DITFEND on English dataset and Chinese 3-domain dataset.}
  \resizebox{2.0\columnwidth}{!}{
    \begin{tabular}{lcccccccccccc}
    \toprule
    \multirow{3}[6]{*}{Method} & \multicolumn{6}{c}{English dataset}            & \multicolumn{6}{c}{Ch-3 dataset} \\
\cline{2-13}          & \multicolumn{3}{c}{target: COVID} & \multicolumn{3}{c}{target: Politifact} & \multicolumn{3}{c}{target: Health} & \multicolumn{3}{c}{target: Politics} \\
\cline{2-13}          & F1    & AUC   & Acc   & F1    & AUC   & Acc   & F1    & AUC   & Acc   & F1    & AUC   & Acc \\
    \cline{1-13}
    BiGRU & 0.7448 & 0.9114 & 0.8606 & 0.7339 & 0.8213 & 0.7375 & 0.8577 & 0.9367 & 0.8580 & 0.8384 & 0.9032 & 0.8637 \\
    \quad +\emph{DITFEND} & \textbf{0.9219} & \textbf{0.9890} & \textbf{0.9501} & \textbf{0.8476} & \textbf{0.9043} & \textbf{0.8477} & \textbf{0.8792} & \textbf{0.9548} & \textbf{0.8795} & \textbf{0.8528} & \textbf{0.9261} & \textbf{0.8772} \\
    \cline{1-13}
    TextCNN & 0.8322 & 0.9397 & 0.8955 & 0.7040 & 0.8046 & 0.7064 & 0.8716 & 0.9552 & 0.8720 & \textbf{0.8579} & 0.9067 & \textbf{0.8859} \\
    \quad +\emph{DITFEND} & \textbf{0.8642} & \textbf{0.9706} & \textbf{0.9184} & \textbf{0.7913} & \textbf{0.8740} & \textbf{0.7913} & \textbf{0.8878} & \textbf{0.9582} & \textbf{0.8880} & 0.8563 & \textbf{0.9127} & 0.8807 \\
    \cline{1-13}
    RoBERTa & 0.9014 & 0.9770 & 0.9377 & 0.7967 & 0.9078 & 0.7989 & 0.8955 & 0.9641 & 0.8955 & 0.8300  & 0.8948 & 0.8628 \\
    \quad +\emph{DITFEND} & \textbf{0.9360} & \textbf{0.9901} & \textbf{0.9578} & \textbf{0.8608} & \textbf{0.9183} & \textbf{0.8609} & \textbf{0.9105} & \textbf{0.9708} & \textbf{0.9105} & \textbf{0.8445} & \textbf{0.9085} & \textbf{0.8725} \\
    \cline{1-13}
    EANN\scriptsize{T}  & 0.8836 & 0.9751 & 0.9282 & 0.7558 & 0.8612 & 0.7584 & 0.9189 & \textbf{0.9787} & 0.9190 & 0.8405 & 0.9074 & 0.8690 \\
    \quad +\emph{DITFEND} & \textbf{0.8883} & \textbf{0.9825} & \textbf{0.9310} & \textbf{0.8040} & \textbf{0.9074} & \textbf{0.8046} & \textbf{0.9219} & 0.9760 & \textbf{0.9220} & \textbf{0.8574} & \textbf{0.9136} & \textbf{0.8854} \\
    \cline{1-13}
    MMOE & \textbf{0.9379} & 0.9883 & 0.9588 & 0.8477  & 0.9408 & 0.8486 & \textbf{0.9215} & 0.9639 & \textbf{0.9215} & \textbf{0.8779} & 0.9388 & \textbf{0.8982} \\
    \quad +\emph{DITFEND} & 0.9361 & \textbf{0.9911} & \textbf{0.9600} & \textbf{0.8613} & \textbf{0.9515} & \textbf{0.8615} & 0.9034 & \textbf{0.9657} & 0.9035 & 0.8523 & \textbf{0.9398} & 0.8766 \\
    \cline{1-13}
    MOSE &0.9326 & 0.9879 & 0.9588 &0.8576 & 0.9447 & 0.8590& 0.9023 & 0.9683 & 0.9025 & 0.8564 & 0.9138 & 0.8795\\
    \quad +\emph{DITFEND} & \textbf{0.9586} & \textbf{0.9880} & \textbf{0.9712} & \textbf{0.8732} & \textbf{0.9553} & \textbf{0.8732} & \textbf{0.9069} & \textbf{0.9711} & \textbf{0.9070} & \textbf{0.8642} & \textbf{0.9220} & \textbf{0.8865}\\
    \cline{1-13}
    EDDFN &0.9306 & 0.9891 & 0.9547 & 0.8505 & 0.9432 & 0.8509& 0.9235 & \textbf{0.9735} & 0.9235 & 0.8440 & 0.9207 & 0.8702 \\
    \quad +\emph{DITFEND} & \textbf{0.9401} & \textbf{0.9912} & \textbf{0.9600} & \textbf{0.8720} & \textbf{0.9466} & \textbf{0.8725} & \textbf{0.9245} & 0.9731 & \textbf{0.9245} & \textbf{0.8486} & \textbf{0.9251} & \textbf{0.8731} \\
    \cline{1-13}
    MDFEND &0.9331  & 0.9874 & 0.9565 & 0.8473 & 0.9391 & 0.8485& 0.9419 & 0.9855 & 0.9420 & 0.8555 & 0.9259 & 0.8854 \\
    \quad +\emph{DITFEND} & \textbf{0.9485} & \textbf{0.9934} & \textbf{0.9700} & \textbf{0.8589} & \textbf{0.9500} & \textbf{0.8593} & \textbf{0.9530} & \textbf{0.9856}  & \textbf{0.9530} &  \textbf{0.8663} & \textbf{0.9368} & \textbf{0.8895} \\
    \bottomrule
    \end{tabular}%
    }
  \label{tab:ench3}%
\end{table*}%

%% file: tables/ablation_ch9.tex
\begin{table}[htbp]
  \centering
  \caption{Ablation study on Chinese 9-domain dataset with RoBERTa and EDDFN.}
  \scalebox{0.66}{
    \begin{tabular}{lcccccc}
    \toprule
     \multirow{2}[4]{*}{Methods} &  \multicolumn{3}{c}{target: Health} & \multicolumn{3}{c}{target: Politics}\\
\cmidrule{2-7}        & F1    & AUC   & Acc  & F1    & AUC   & Acc  \\
    \midrule
    RoBERTa & 0.9090 & 0.9611 & 0.9090  &  0.8366  & 0.9034  &  0.8637 \\
    \quad +\emph{DITFEND} & \textbf{0.9115} & \textbf{0.9739} & \textbf{0.9115} &   \textbf{0.8775}    &   \textbf{0.9199}    &  \textbf{0.8982}\\
          \quad \emph{w/o} meta  & 0.9096   &   0.9654    &  0.9087 & 0.8557 & 0.9144 & 0.8795\\
          \quad \emph{w/o} sources  & 0.9100    &   0.9712    & 0.9100 & 0.8687 & 0.9183 & 0.8928\\
    \midrule
    EDDFN &  0.9379 & 0.9807 & 0.9380  & 0.8478 & 0.9292 & 0.8754\\
    \quad +\emph{DITFEND} &   \textbf{0.9399} & \textbf{0.9821} & \textbf{0.9400} & \textbf{0.8507} & \textbf{0.9308} & 0.8772\\
          \quad \emph{w/o} meta &    0.9380   &   0.9810    &  0.9380 & 0.8480 & 0.9295 & 0.8760\\
          \quad \emph{w/o} sources &    0.9385    &   0.9815    & 0.9383 & 0.8482 & 0.9302 & \textbf{0.8798}\\
    \bottomrule
    \end{tabular}%
     }   
  \label{tab:abla_ch9}%
\end{table}%

%% file: tables/pp_compare.tex
\begin{table}[htbp]
  \centering
  \footnotesize
  \caption{Some representative examples selected depending on Transferability Quantifier. Words or phrases underlined are some indicators related to the target domain.}
    \begin{tabular}{p{24.00em}}
    \toprule
    \rowcolor[rgb]{ .851,  .851,  .851} \multicolumn{1}{l}{Target: Health} \\
    \midrule
    I. The team will undergo \underline{nucleic acid tests} in the coming days as they resume individual training. [entertainment]
    \newline{}II. No one was \underline{injured} when a test carriage derailed. [disaster]
    \newline{}III. Bask in inflatable dolls caused police cars and \underline{ambulances} all out in broad daylight. [society]\\
    \midrule
    \rowcolor[rgb]{ .851,  .851,  .851} \multicolumn{1}{l}{Target: Finance} \\
    \midrule
    I. It has become the first \underline{Tencent smart medical joint} \underline{innovation base} in China.  [health]
    \newline{}II. \underline{The central bank} claimed to accelerate finance-technology regulatory framework. [science]
    \newline{}III. There are barriers to \underline{electronic use} for older people in digital lifestyle. [society]\\
    \midrule
    \rowcolor[rgb]{ .851,  .851,  .851} \multicolumn{1}{l}{Target: Politics} \\
    \midrule
    I. \underline{Trump adviser} urges Taiwan to come forward with advanced weapons program.  [military]
    \newline{}II. \underline{Shandong provincial government} and Tencent signed a strategic cooperation agreement. [Finance]
    \newline{}III. Beijing municipal \underline{government} announces the hukou score to the society every year. [society]\\
    \bottomrule
    \end{tabular}%
  \label{tab:case}%
\end{table}%

%% file: tables/online_result.tex

\begin{table}[htbp]
  \centering
  \small
  \caption{Online test performance. The results below are a relative improvement over the two baseline models.}
    \begin{tabular}{lcc}
    \toprule
    Improvement on & SPAUC & AUC \\
    \midrule
    TextCNN & +2.07\% & +1.40\% \\
    EANN  & +3.40\% & +2.90\% \\
    \bottomrule
    \end{tabular}%
  \label{tab:online}%
\end{table}%

%% file: Conclusion.tex
\section{Conclusion and Future Work}
\textbf{Conclusion.} In this paper, we propose DITFEND, a Domain- and Instance-level Transfer framework to improve fake news detection in an influential domain. For domain-level transfer, we adopt meta-learning to learn common knowledge across domains, in which way we can learn a general model; For instance-level transfer, we transform the problem of transferability evaluation to the task of character/word prediction by a language model endowed with target knowledge. Experiments on the Chinese dataset and English dataset demonstrate the effectiveness of our DITFEND framework over several fake news detection models. Ablation studies and case studies further evaluate the effectiveness and  interpretability of each module in DITFEND. Online testing results verified the practical performance of our DITFEND framework.

\textbf{Future Work.} (1) In this work, the domain label of the target domain is unknown in advance. We will investigate how to handle the situation when the domain label of the target domain and/ or source domains is unknown; (2) Temporal distribution shift is more challenging, and we plan to investigate how to adapt fake news detection models to the data in the future; (3) We will explore how to select source domains based on the analysis of transferability between different domains.

%% file: Appendix.tex
\section{Supplementary Experiments}
\label{sec:appendix}
\input{appendix_table/ch6_ch9}

\subsection{Experiments on Chinese 6-domain and Chinese 9-domain Dataset}
\label{sec: ch69experiments}
Experimental results on the Chinese 6-domain and Chinese 9-domain datasets are listed in Table ~\ref{tab:ch6_ch9}.

\subsection{Experiments on Other Target Domains}
In this section, we set other domains, which have not been chosen as target domains in Section ~\ref{sec:experiment}, and evaluate the performance of the proposed DITFEND on these domains. Experimental results are shown in Table ~\ref{tab:add_target}.
\input{appendix_table/add_target}
\subsection{Experiments on New Target Domains}
In this section, we perform experiments to exclude target domain data when training the general model, in order to testify how our proposed framework performs when the target domain is unseen during general model training procedure. 

We show the experimental results in Table \ref{tab:target_unseen_en_ch3} and Table \ref{tab:target_unseen_ch6_ch9}.
From the experimental results, we can find that our framework DITFEND can bring significant improvements to all domains. 

\input{appendix_table/target_unseen_en_ch3}
\input{appendix_table/target_unseen_ch6_ch9}

%% file: appendix_table/ch6_ch9.tex
\begin{table*}[htbp]
  \centering
  \caption{Performance comparison of base models with and without DITFEND on Chinese 6-domain dataset and Chinese 9-domain dataset. }
  \resizebox{2.0\columnwidth}{!}{
    \begin{tabular}{lcccccccccccc}
    \toprule
    \multirow{3}[6]{*}{Method} & \multicolumn{6}{c}{Ch-6 dataset}            & \multicolumn{6}{c}{Ch-9 dataset} \\
\cline{2-13}          & \multicolumn{3}{c}{target: Health} & \multicolumn{3}{c}{target: Finance} & \multicolumn{3}{c}{target: Health} & \multicolumn{3}{c}{target: Politics} \\
\cline{2-13}          & F1    & AUC   & Acc   & F1    & AUC   & Acc   & F1    & AUC   & Acc   & F1    & AUC   & Acc \\
    \cline{1-13}
    BiGRU & 0.8626 & 0.9442 & 0.8630 &   0.8254    &   0.9281    &   0.8642    & \textbf{0.8868} & 0.9574 & \textbf{0.8870} & 0.8356 & 0.9119 & 0.8590 \\
    \quad +\emph{DITFEND} & \textbf{0.8991} & \textbf{0.9676} & \textbf{0.8992} &   \textbf{0.8431}    &   \textbf{0.9343}    &   \textbf{0.8755}    & 0.8789 & \textbf{0.9630} & 0.8794 & \textbf{0.8749} & \textbf{0.9334} & \textbf{0.8925} \\
    \cline{1-13}
    TextCNN & 0.8832 & \textbf{0.9670} & 0.8835 &  \textbf{0.8646}     &   \textbf{0.9498}    &   \textbf{0.8943}    & 0.8768 & 0.9556 & 0.8770 & 0.8561 & 0.9225 & 0.8813 \\
    \quad +\emph{DITFEND} & \textbf{0.8922} & 0.9631 & \textbf{0.8925} &   0.8581    &  0.9436     &    0.8887    & \textbf{0.9093} & \textbf{0.9742} & \textbf{0.9094} & \textbf{0.8768} & \textbf{0.9235} & \textbf{0.8962} \\
    \cline{1-13}
    RoBERTa & 0.9100  & 0.9644 & 0.9100  &  0.8700    &   \textbf{0.9553}    &  0.8989    & 0.9090 & 0.9611 & 0.9090 & 0.8366 & 0.9034 & 0.8637 \\
    \quad +\emph{DITFEND} & \textbf{0.9175} & \textbf{0.9769} & \textbf{0.9175} &  \textbf{0.8768}     &   0.9538    &   \textbf{0.9044}    & \textbf{0.9115} & \textbf{0.9739} & \textbf{0.9115} & \textbf{0.8775} & \textbf{0.9199} & \textbf{0.8982} \\
    \cline{1-13}
    EANN  & 0.9150 & 0.9761 & 0.9150 &   \textbf{0.8621}    &   \textbf{0.9483}    &   0.8906    & 0.9150 & 0.9762 & 0.9150 & 0.8705 & 0.9176 & 0.8918 \\
    \quad +\emph{DITFEND} & \textbf{0.9224} & \textbf{0.9767} & \textbf{0.9225} &   0.8538    &   0.9466    &   \textbf{0.8918}    & \textbf{0.9250} & \textbf{0.9797} & \textbf{0.9250}  & \textbf{0.8822} & \textbf{0.9367} & \textbf{0.9035} \\
    \cline{1-13}
    MMOE & 0.9260 & 0.9754 & 0.9260 & 0.8546 & 0.9501 & 0.8887  & \textbf{0.9364}  & \textbf{0.9774} & \textbf{0.9365} & 0.8620 & 0.9314 & 0.8842\\
    \quad +\emph{DITFEND} & \textbf{0.9271} & \textbf{0.9756} & \textbf{0.9271} & \textbf{0.8611} & \textbf{0.9563} & \textbf{0.8906} & 0.9199 & 0.9655 & 0.9200 & \textbf{0.8633} & \textbf{0.9357} & \textbf{0.8860} \\
    \cline{1-13}
    MOSE & 0.9118 & 0.9720 & 0.9120 & \textbf{0.8639} & 0.9500 & \textbf{0.8921} & \textbf{0.9179} & 0.9700 & \textbf{0.9160} & 0.8673 & 0.9388 & 0.8918\\
    \quad +\emph{DITFEND} & \textbf{0.9200} & \textbf{0.9735} & \textbf{0.9200} &  0.8502& \textbf{0.9533} & 0.8830 & 0.9050 & \textbf{0.9725} & 0.9050 & \textbf{0.8681} & \textbf{0.9422} & \textbf{0.8972} \\
    \cline{1-13}
    EDDFN & 0.9280 & 0.9774 & 0.9280 & 0.8456 & 0.9436 & 0.8830 & 0.9379 & 0.9807 & 0.9380  & 0.8478 & 0.9292 & 0.8754\\
    \quad +\emph{DITFEND} & \textbf{0.9350} & \textbf{0.9821} & \textbf{0.9350} & \textbf{0.8801} & \textbf{0.9434} & \textbf{0.9019} & \textbf{0.9399} & \textbf{0.9821} & \textbf{0.9400} & \textbf{0.8507} & \textbf{0.9308} & \textbf{0.8772}\\
    \cline{1-13}
    MDFEND & 0.9430 & 0.9851 & 0.9430 & 0.8749 & 0.9610 & 0.9023 & 0.9425 & 0.9846 & 0.9425& 0.8774 & 0.9370 & 0.8994  \\
    \quad +\emph{DITFEND} & \textbf{0.9451} & \textbf{0.9876} & \textbf{0.9450} & \textbf{0.8800} & \textbf{0.9649} & \textbf{0.9100} & \textbf{0.9500} & \textbf{0.9864} & \textbf{0.9500} & \textbf{0.8986} & \textbf{0.9541} & \textbf{0.9181} \\
    \bottomrule
    \end{tabular}%
    }
  \label{tab:ch6_ch9}%
\end{table*}%

%% file: appendix_table/add_target.tex
\begin{table*}[htbp]
  \centering
  \caption{Performance comparison (f1-score) of base models with and without DITFEND when assigning other target domains on Chinese 9-domain dataset.}
    \label{tab:add_target}%
    \resizebox{1.6\columnwidth}{!}{
    \begin{tabular}{lrrrrrrrrr}
    \toprule
    \multicolumn{1}{l}{target domain} & \multicolumn{1}{c}{science} & \multicolumn{1}{c}{military} & \multicolumn{1}{c}{Edu.} & \multicolumn{1}{c}{Disaster} & \multicolumn{1}{c}{Politics} & \multicolumn{1}{c}{Health} & \multicolumn{1}{c}{Finance} & \multicolumn{1}{c}{Ent.} & \multicolumn{1}{c}{Society} \\
    \cline{1-10}
    BiGRU & 0.7269  & 0.8724  & 0.8138  & 0.7935  & 0.8356  & \textbf{0.8868}  & 0.8291  & 0.8629  & \textbf{0.8485}\\
    \quad +\emph{DITFEND} &   \textbf{0.8061}    &   \textbf{0.8604}    &   \textbf{0.9089}    &  \textbf{0.8017}     &   \textbf{0.8749}    &    0.8789   &    \textbf{0.8665}   &   \textbf{0.8840}    & 0.8483 \\
    \cline{1-10}
    TextCNN & 0.7254  & 0.8839  & 0.8362  & 0.8222  & 0.8561  & 0.8768  & 0.8638  & 0.8456  & 0.8540\\
    \quad +\emph{DITFEND} &   \textbf{0.7283}    &   \textbf{0.8882}    &   \textbf{0.8829}    &   \textbf{0.8301}    &   \textbf{0.8768}    &   \textbf{0.9093}    &   \textbf{0.8665}    &   \textbf{0.8840}    & \textbf{0.8769} \\
    \cline{1-10}
    RoBERTa & 0.7777  & 0.9072  & 0.8331  & 0.8512  & 0.8366  & 0.9090  & 0.8735  & 0.8769  & 0.8577\\
    \quad +\emph{DITFEND} &   \textbf{0.8107}    &   \textbf{0.9129}    &   \textbf{0.8384}   &  \textbf{0.8574}     &   \textbf{0.8775}    &   \textbf{0.9115}    &   \textbf{0.8889}    &   \textbf{0.8889}    &  \textbf{0.8881} \\
    \cline{1-10}
    EANN\scriptsize{T}  & 0.8225  & 0.9274  & \textbf{0.8624}  & 0.8666  & 0.8705  & 0.9150  & 0.8710  & 0.8957  & \textbf{0.8877}\\
    \quad +\emph{DITFEND} &   \textbf{0.8271}    &   \textbf{0.9419}    &   0.8582    & \textbf{0.9060}      &   \textbf{0.8833}    &   \textbf{0.9250}    &   \textbf{0.8833}    &   \textbf{0.9100}    &  0.8874\\
    \cline{1-10}
    MMOE  & \textbf{0.8755}  & 0.9112  & 0.8706  & 0.8770  & 0.8620  & \textbf{0.9364}  & 0.8567  & 0.8886  & 0.8750\\
    \quad +\emph{DITFEND} &   0.8587    &   \textbf{0.9186}    &    \textbf{0.8769}   &  \textbf{0.8836}     &   \textbf{0.8792}    &   0.9350    &   \textbf{0.8786}    &   \textbf{0.8923}    & \textbf{0.8895} \\
    \cline{1-10}
    MOSE  &  0.8502  & 0.8858  & 0.8815  & \textbf{0.8672}  & 0.8808  & 0.9179  & 0.8672  & 0.8913  & 0.8729\\
    \quad +\emph{DITFEND} &    \textbf{0.8656}   &    \textbf{0.9121}   &   \textbf{0.8918}    &  0.8627     &   \textbf{0.8897}    &    \textbf{0.9200}   &   \textbf{0.8846}    &   \textbf{0.8987}    &  \textbf{0.8832}  \\
    \cline{1-10}
    EDDFN  & \textbf{0.8186}  & \textbf{0.9137}  & 0.8676  & 0.8786  & 0.8478  & 0.9379  & 0.8636  & 0.8832  & 0.8689  \\
    \quad +\emph{DITFEND} &   0.8061    &   0.8834    &   \textbf{0.8887}    &   \textbf{0.8855}    &   \textbf{0.8492}    &   \textbf{0.9399}    &   \textbf{0.8647}    &   \textbf{0.8889}   &  \textbf{0.8754}  \\
    \cline{1-10}
    MDFEND  & 0.8301  & 0.9389  & 0.8917  & 0.9003  & 0.8865  & 0.9400  & 0.8951  & 0.9066  & 0.8980 \\
    \quad +\emph{DITFEND} &   \textbf{0.8426}    &   \textbf{0.9419}    &   \textbf{0.9091}    &  \textbf{0.9145}     &    \textbf{0.8895}   &   \textbf{0.9550}    &   \textbf{0.8984}    &   \textbf{0.9100}    &  \textbf{0.8991} \\
    \hline
    \end{tabular}%
    }
\end{table*}%

%% file: appendix_table/target_unseen_en_ch3.tex
\begin{table*}[htbp]
  \centering
  \caption{Performance comparison of base models with and without DITFEND on English dataset and Chinese 3-domain dataset when the target domain is unseen during meta-training procedure. }
  \resizebox{2.0\columnwidth}{!}{
    \begin{tabular}{lcccccccccccc}
    \toprule
    \multirow{3}[6]{*}{Method} & \multicolumn{6}{c}{English dataset}            & \multicolumn{6}{c}{Ch-3 dataset} \\
\cline{2-13}          & \multicolumn{3}{c}{target: COVID} & \multicolumn{3}{c}{target: Politifact} & \multicolumn{3}{c}{target: Health} & \multicolumn{3}{c}{target: Politics} \\
\cline{2-13}          & F1    & AUC   & Acc   & F1    & AUC   & Acc   & F1    & AUC   & Acc   & F1    & AUC   & Acc \\
    \cline{1-13}
    BiGRU & 0.7448 & 0.9114 & 0.8606 & 0.7339 & 0.8213 & 0.7375 & 0.8577 & 0.9367 & 0.8580 & 0.8384 & 0.9032 & 0.8637 \\
    \quad +\emph{DITFEND} & \textbf{0.8949} & \textbf{0.9650} & \textbf{0.8950} & \textbf{0.8406} & \textbf{0.9052} & \textbf{0.8406} & \textbf{0.8746} & \textbf{0.9609} & \textbf{0.8750} & \textbf{0.8633} & \textbf{0.9316} & \textbf{0.8889} \\
    \cline{1-13}
    TextCNN & 0.8322 & 0.9397 & 0.8955 & 0.7040 & 0.8046 & 0.7064 & 0.8716 & 0.9552 & 0.8720 & 0.8579 & 0.9067 & \textbf{0.8859} \\
    \quad +\emph{DITFEND} & \textbf{0.8549} & \textbf{0.9658} & \textbf{0.9042} &\textbf{0.7851} & \textbf{0.8632} & \textbf{0.7850} & \textbf{0.8949} & \textbf{0.9650} & \textbf{0.8950} & \textbf{0.8587} & \textbf{0.9615} & 0.8810 \\
    \cline{1-13}
    RoBERTa & 0.9014 & 0.9770 & 0.9377 & 0.7967 & 0.9078 & 0.7989 & 0.8955 & 0.9641 & 0.8955 & 0.8300  & 0.8948 & 0.8628 \\
    \quad +\emph{DITFEND} & \textbf{0.9305} & \textbf{0.9800} & \textbf{0.9509} & \textbf{0.8493} & \textbf{0.9335} & \textbf{0.8497}& \textbf{0.9250} & \textbf{0.9743} & \textbf{0.9250} & \textbf{0.8302} & \textbf{0.9027} & \textbf{0.8655} \\
    \cline{1-13}
    EANN\scriptsize{T}  & 0.8836 & 0.9751 & 0.9282 & 0.7558 & 0.8612 & 0.7584 & 0.9189 & \textbf{0.9787} & 0.9190 & 0.8405 & 0.9074 & 0.8690 \\
    \quad +\emph{DITFEND} & \textbf{0.9507} &\textbf{0.9896} & \textbf{0.9667} & \textbf{0.8664} & \textbf{0.9597} & \textbf{0.8671} & \textbf{0.9200} & 0.9768 & \textbf{0.9200} & \textbf{0.8615} & \textbf{0.9177} & \textbf{0.8889} \\
    \cline{1-13}
    MMOE & 0.9379 & 0.9883 & \textbf{0.9588} & 0.8477  & 0.9408 & 0.8486 & 0.9215 & 0.9679 & 0.9215 & \textbf{0.8779} & \textbf{0.9388} & \textbf{0.8982}\\
    \quad +\emph{DITFEND} & \textbf{0.9356} & \textbf{0.9907} & 0.9586 & \textbf{0.8542} & \textbf{0.9482} & \textbf{0.8555} & \textbf{0.9299} & \textbf{0.9735} & \textbf{0.9300} & 0.8659 & 0.9331 & 0.8896\\
    \cline{1-13}
    MOSE & 0.9326 & 0.9879 & 0.9588 &0.8576 & 0.9447 & 0.8590 & \textbf{0.9023} & 0.9683 & \textbf{0.9025} & 0.8564 & \textbf{0.9138} & 0.8795\\
    \quad +\emph{DITFEND} & \textbf{0.9547} & \textbf{0.9889} & \textbf{0.9692} & \textbf{0.8725} & \textbf{0.9528} & \textbf{0.8728} & 0.9010 & \textbf{0.9685} & 0.9012 & \textbf{0.8568} & 0.9137 & \textbf{0.8812} \\
    \cline{1-13}
    EDDFN & 0.9306 & 0.9891 & 0.9547 & 0.8505 & 0.9432 & 0.8509 & \textbf{0.9235} & 0.9735 & \textbf{0.9235} & 0.8440 & 0.9207 & \textbf{0.8702} \\
    \quad +\emph{DITFEND} & \textbf{0.9383} & \textbf{0.9899} & \textbf{0.9586} & \textbf{0.8721} & \textbf{0.9466} & \textbf{0.8728} & 0.9150 & \textbf{0.9762} & 0.9150 & \textbf{0.8441} & \textbf{0.9337} & 0.8663 \\
    \cline{1-13}
    MDFEND & 0.9331  & 0.9874 & 0.9565 & 0.8473 & 0.9391 & 0.8485 & 0.9419 & 0.9855 & 0.9420 & \textbf{0.8555} & 0.9259 & 0.8854 \\
    \quad +\emph{DITFEND} & \textbf{0.9482} & \textbf{0.9932} & \textbf{0.9659} & \textbf{0.8599} & \textbf{0.9503} & \textbf{0.8613} & \textbf{0.9550} & \textbf{0.9877} & \textbf{0.9550} & 0.8507 & \textbf{0.9442} & \textbf{0.8772}  \\
    \bottomrule
    \end{tabular}%
    }
  \label{tab:target_unseen_en_ch3}%
\end{table*}%

%% file: appendix_table/target_unseen_ch6_ch9.tex
\begin{table*}[htbp]
  \centering
  \caption{Performance comparison of base models with and without DITFEND on Chinese 6-domain dataset and Chinese 9-domain dataset when the target domain is unseen during meta-training procedure.}
  \resizebox{2.0\columnwidth}{!}{
    \begin{tabular}{lcccccccccccc}
    \toprule
    \multirow{3}[6]{*}{Method} & \multicolumn{6}{c}{Ch-6 dataset}            & \multicolumn{6}{c}{Ch-9 dataset} \\
\cline{2-13}          & \multicolumn{3}{c}{target: Health} & \multicolumn{3}{c}{target: Finance} & \multicolumn{3}{c}{target: Health} & \multicolumn{3}{c}{target: Politics} \\
\cline{2-13}          & F1    & AUC   & Acc   & F1    & AUC   & Acc   & F1    & AUC   & Acc   & F1    & AUC   & Acc \\
    \cline{1-13}
    BiGRU & 0.8626 & 0.9442 & 0.8630 &   0.8254    &   0.9281    &   0.8642    & \textbf{0.8868} & 0.9574 & \textbf{0.8870} & 0.8356 & 0.9119 & 0.8590 \\
    \quad +\emph{DITFEND} & \textbf{0.8875} & \textbf{0.9548} & \textbf{0.8875} & \textbf{0.8400} & \textbf{0.9295} & \textbf{0.8710} & 0.8700 & \textbf{0.9612} & 0.8715 & \textbf{0.8541} & \textbf{0.9305} & \textbf{0.8816}\\
    \cline{1-13}
    TextCNN & 0.8832 & \textbf{0.9670} & 0.8835 &  \textbf{0.8646}     &   \textbf{0.9498}    &   \textbf{0.8943}    & 0.8768 & 0.9556 & 0.8770 & 0.8561 & 0.9225 & 0.8813 \\
    \quad +\emph{DITFEND} & \textbf{0.8901} & 0.9608 & \textbf{0.8902} & 0.8547 & 0.9431 & 0.8814 & \textbf{0.8854} & \textbf{0.9715} & \textbf{0.8855} & \textbf{0.8650} & \textbf{0.9230} & \textbf{0.8879}\\
    \cline{1-13}
    RoBERTa & 0.9100  & 0.9644 & 0.9100  &  0.8700    &   \textbf{0.9553}    &  0.8989    & 0.9090 & 0.9611 & 0.9090 & 0.8366 & 0.9034 & 0.8637 \\
    \quad +\emph{DITFEND} & \textbf{0.9249} & \textbf{0.9666} & \textbf{0.9250} & \textbf{0.8781} & 0.9347 & \textbf{0.9057} & \textbf{0.9149} & \textbf{0.9723} & \textbf{0.9150} & \textbf{0.8999} & \textbf{0.9294} & \textbf{0.9181} \\
    \cline{1-13}
    EANN\scriptsize{T}  & 0.9150 & 0.9761 & 0.9150 &   0.8621    &   0.9483    &   0.8906    & 0.9150 & \textbf{0.9762} & 0.9150 & 0.8705 & 0.9176 & 0.8918 \\
    \quad +\emph{DITFEND} & \textbf{0.9184} & \textbf{0.9753} & \textbf{0.9184} & \textbf{0.8789} & \textbf{0.9649} & \textbf{0.9094} & \textbf{0.9196} & 0.9676 & \textbf{0.9196} & \textbf{0.8733} & \textbf{0.9178} & \textbf{0.8940}\\
    \cline{1-13}
    MMOE & \textbf{0.9260} & \textbf{0.9754} & \textbf{0.9260} & 0.8546 & 0.9501 & 0.8887& \textbf{0.9364}  & \textbf{0.9774} & \textbf{0.9365} & \textbf{0.8620} & 0.9314 & 0.8842\\
    \quad +\emph{DITFEND} & 0.9145 & 0.9732 & 0.9150 & \textbf{0.8548} & \textbf{0.9503} & \textbf{0.8888} & 0.9200 & 0.9663 & 0.9200 & 0.8496 & \textbf{0.9459} & \textbf{0.8843} \\
    \cline{1-13}
    MOSE & \textbf{0.9118} & 0.9720 & \textbf{0.9120} & 0.8639 & 0.9500 & 0.8921&  \textbf{0.9179} & 0.9700 & \textbf{0.9160} & \textbf{0.8673} & \textbf{0.9388} & \textbf{0.8918}\\
    \quad +\emph{DITFEND} & 0.9110 & \textbf{0.9725} & 0.9112 & \textbf{0.8640} & \textbf{0.9512} & \textbf{0.8925} & 0.9048 & \textbf{0.9715} & 0.9049 & 0.8540 & 0.9386 & 0.8713\\
    \cline{1-13}
    EDDFN &0.9280 & \textbf{0.9774} & 0.9280 & 0.8456 & 0.9436 & 0.8830& 0.9379 & \textbf{0.9807} & 0.9380  & \textbf{0.8478} & \textbf{0.9292} & \textbf{0.8754} \\
    \quad +\emph{DITFEND} & \textbf{0.9299} & 0.9745 & \textbf{0.9300} & \textbf{0.8471} & \textbf{0.9497} & \textbf{0.8868} & \textbf{0.9399} & 0.9768 & \textbf{0.9400} & 0.8240 & 0.8951 & 0.8596 \\
    \cline{1-13}
    MDFEND &0.9430 & 0.9851 & 0.9430 & 0.8749 & \textbf{0.9610} & 0.9023 & 0.9425 & 0.9846 & 0.9425& \textbf{0.8774} & 0.9370 & \textbf{0.8994} \\
    \quad +\emph{DITFEND} & \textbf{0.9450} & \textbf{0.9900} & \textbf{0.9450} & \textbf{0.8792} & 0.9486 & \textbf{0.9057} & \textbf{0.9600} & \textbf{0.9892} & \textbf{0.9600} &0.8713 & \textbf{0.9452} & 0.8947 \\
    \bottomrule
    \end{tabular}%
    }
  \label{tab:target_unseen_ch6_ch9}%
\end{table*}%